\documentclass{article}
\usepackage{ijcai21}

\usepackage{times}

\usepackage{soul}
\usepackage{url}
\usepackage[hidelinks]{hyperref}
\usepackage[utf8]{inputenc}
\usepackage[small]{caption}
\usepackage{graphicx}
\usepackage{amsmath}
\usepackage{booktabs}
\title{Pattern Transfer Learning for Reinforcement Learning in Order Dispatching} %  Reinforcement Learning via Concordance Penalty}

\author{
Runzhe Wan $^{1}$\footnote{Equal contribution}\and
Sheng Zhang$^{1*}$\and
Chengchun Shi$^{2}$\and
Shikai Luo$^3$\And
Rui Song$^1$\\
\affiliations
$^1$ North Carolina State University\\
$^2$ London School of Economics and Political Science\\
$^3$ Tecent PCG\\
\emails
\{rwan, szhang37\}@ncsu.edu,
c.shi7@lse.ac.uk,
sluo198912@163.com,
rsong@ncsu.edu
}

\newcommand{\I}{\mathbb{I}}

\usepackage{soul}
\usepackage{xcolor}

\newcommand{\vV}{\boldsymbol{V}}

\newtheorem{proposition}{Proposition}

\DeclareMathOperator*{\argmax}{arg\,max}
\DeclareMathOperator*{\argmin}{arg\,min}

\usepackage{microtype}
\usepackage{graphicx}

\usepackage{booktabs} % for professional tables
\usepackage{amsmath}
\usepackage{amssymb}
\usepackage{hyperref}
\usepackage{bm}
\usepackage{soul}

\usepackage{xr}

\usepackage{graphicx}

\usepackage[utf8]{inputenc} % allow utf-8 input
\usepackage[T1]{fontenc}    % use 8-bit T1 fonts
\usepackage{hyperref}       % hyperlinks
\usepackage{url}            % simple URL typesetting
\usepackage{booktabs}       % professional-quality tables
\usepackage{amsfonts}       % blackboard math symbols
\usepackage{nicefrac}       % compact symbols for 1/2, etc.
\usepackage{microtype}      % microtypography
\usepackage{lipsum}		% Can be removed after putting your text content
\usepackage{microtype}
\usepackage{graphicx}
\usepackage{booktabs} % for professional tables
\usepackage{amsmath}
\usepackage{amssymb}
\usepackage{natbib}

\usepackage{hyperref}
\usepackage{bm}
\usepackage{xr}
\usepackage{diagbox}
\usepackage{setspace}
\usepackage{subcaption}
\RequirePackage{fancyhdr}
\RequirePackage{color}
\RequirePackage{algorithm}
\RequirePackage{algorithmic}
\RequirePackage{natbib}
\RequirePackage{eso-pic} % used by \AddToShipoutPicture 
\RequirePackage{forloop}
\usepackage{sectsty}

\begin{document}

\maketitle

\let\thefootnote\relax\footnotetext{\\Reinforcement Learning for Intelligent Transportation Systems (RL4ITS) Workshop in 
\textit{the $30^{th}$ International
Joint Conference on Artificial Intelligence (IJCAI-21)
}, Montrealthemed Virtual Reality, Canada, 2021. Copyright 2021 by the author(s).}

\begin{abstract}
Order dispatch is one of the central problems to ride-sharing platforms. 
Recently, value-based reinforcement learning algorithms have shown promising performance  to solve this task. 
% under stationarity setting of environment.
% With the development of Markov Decision Process (MDP) framework for the order-dispatch problem, existed reinforcement learning  based methods have achieved significant 
However, in real-world applications, 
the demand-supply system is typically nonstationary over time, posing challenges to re-utilizing data generated in different time periods to learn the value function. 
% the non-stationarity of the demand-supply system poses challenges to 
% Effectively transferring value patterns learned in a source environment to a target environment is critical to improving the performance of an order dispatch algorithm. 
% Concordance, which measures the agreement between two variables, e.g. to evaluate hot zones in different environment, provides a good resource to address the challenge. 
In this work, motivated by the fact that the relative relationship between the values of some states is largely stable across various environments, we propose a pattern transfer learning framework for value-based reinforcement learning in the order dispatch problem. 
%At the heart of our method is a concordance penalty, which efficiently captures the value patterns. 
Our method efficiently captures the value patterns by incorporating a concordance penalty.
The superior performance of the proposed method is supported by experiments. 
\end{abstract}

\section{Introduction}

% 介绍 为什么要做 order dispatch

% For ride-sourcing platforms, such as Uber and DiDi Chuxing, numerous spatio-temporal demand and supply data are collected everyday. 
One major task for large-scale ride-sourcing platforms, such as Uber and DiDi Chuxing, is to develop an order dispatch algorithm which matches order requests with idle drivers in real time. 
% and on a large scale, with the objective of maximizing the long-term reward. 
A high-quality dispatch algorithm can alleviate the traffic congestion problem, increase revenue for drivers, and serve customers better with higher answer rates \citep{xu2018}. 
% Therefore, how to improve the current order dispatch efficiency becomes an active applied research topic. 
% 介绍 别人做了啥, 问题是啥

In recent years, value-based reinforcement learning (RL) algorithms have been widely used in the order dispatch problem  \citep{tang2019deep, zhou2021graph}.  
% One major challenge to these algorithms is how 
% to accurately estimate the value functions with limited data. 
One major challenge to these algorithms is how to accurately estimate the value functions  with limited real-time data.
Although there are usually lots of logged historical data in ride-sourcing platforms, 
given the non-stationarity of the environment, the complex spatial-temporal dependency of this problem, and the multi-agent nature of this task, 
it is not clear how to re-utilize historical data generated by a different policy in a different time period, in a principled way \citep{qin2020ride}.
Naively combining all data sources may cause huge bias, while simply discarding historical data may cause large variance in value estimation and hence affect the dispatch quality. 
Most existing methods fail to address this important problem. 

% performance of the resulting dispatch algorithm.  
% Therefore, the current practice only uses the data accumulated during the recent several days to estimate the value functions \citep{qin2020ride} and discards the data in previous period. 

% to update the policy every several days and estimate the value of the new policy only using data accumulated during the several days \citep{qin2020ride}. 

\begin{figure}[h]
%   \centering
\hspace{-.5cm}
  \includegraphics[width=0.5\textwidth]{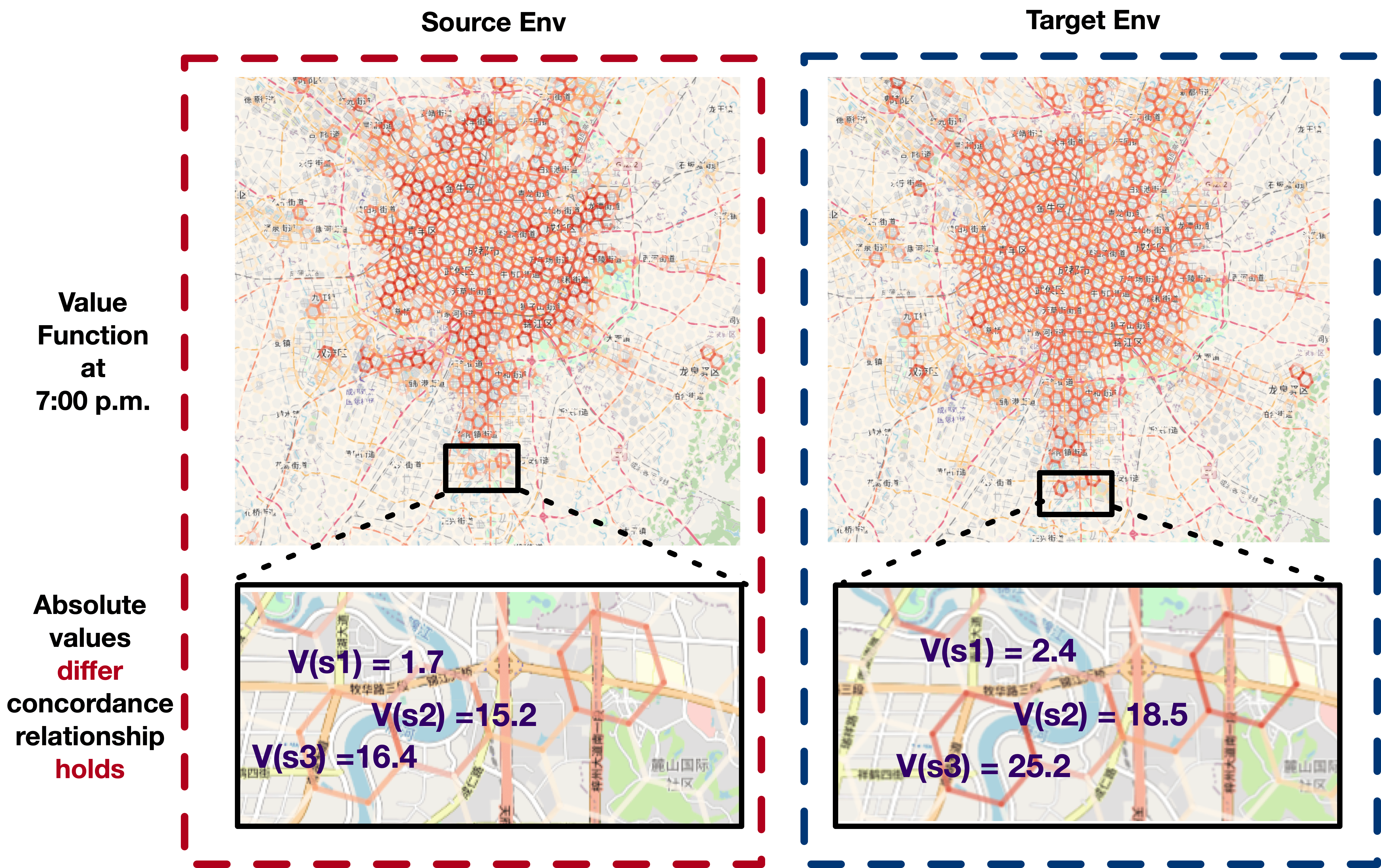}
 \caption{State values over two different time periods. 
 Left and right panels denotes source and target environment, respectively. 
 Computed with the data and setting introduced in Section \ref{sec:exp} using Equation \eqref{eqn:DP_value}. 
 A darker color indicates a higher value.
 The second row displays a zoom-in area, where it is clear that the absolute values in different time periods significantly differ while the relative relationship between states is consistent, which is referred to as the concordance relationship. 
 Overall, the concordance relationship holds on more than 80\% state pairs.}
 \label{fig:data}
\end{figure}
% Q: but absolute value differs?

% 我们打算咋改进 --> 利用某些assumption (基于某些假设) --> 然后用concordance penalty 来估计 value
This paper is concerned with the following question: 
\textit{how should we utilize  data generated in a different environment to improve the efficiency of value function estimation in a new environment?} 
Such a goal may sound ambitious, and certain task-specific structures must be utilized. 
% In this work, we investigate the role of structural penalty functions in pattern transfer. 
% to transfer pattern in source environment to target environment.
% penalty
In this work, we focus on the concordance pattern, motivated by the following observations: 
in a demand-supply network, although the absolute value of a  state may vary a lot across different environments, the relative relationship between some states tends to be stationary. 
For example, as shown in Figure \ref{fig:data}, there exist some "hot zones" in the center of the town, which have higher values than the values of the "cold zones" in the suburbs.
This relationship is stationary across different periods, and can be characterized by the concordance relationship between the value functions in different environments. 
Such a relationship also exists between, for example, the value of the rush hours and the normal hours. 
Therefore, such a pattern relationship can then be used to improve the efficiency of value estimation, 
by appropriately integrating a concordance penalty with a value-based RL algorithm. 
% via a concordance penalty, after appropriately integrating the penalty with a value-based RL algorithm. 
% In other words, although the absolute value of the value function  may vary significantly, the relative relationship of values may be consistent, across different time/units/locations, etc. 

% Therefore, we conduct the value function estimation step with concordance constraint. 
% The details about value evaluation can be found in Section 4.2.  

% , and the corresponding concordance penalty can help stabilize value estimation. 

% 总结
% The proposed method utilizes the pattern of the value function  learned in a source environment to boost the training process in our target environment.  
% To the best of our knowledge, we are the first to integrate concordance penalty into the value evaluation of order dispatch to provide the pattern knowledge for different time periods. In summary, :

%  capturing the pattern  consistency between environments via the concordance relationship, and stablize the value estimation with a concordance penalty. 
In this paper, we investigate pattern transfer learning for the order dispatch problem. 
Our contributions are three-fold: 
\begin{itemize}
    \item Conceptually, we demonstrate a concrete approach to transfer the structural information learned from existing offline data to improve the efficiency of online value estimation. % in a different environment
    The high-level idea of utilizing a structural penalty function in transfer RL is generally applicable and is new to the literature. 
    % , and to our knowledge,  this is the first work  by using a structural penalty. 
    \item Methodologically, we develop a novel transfer RL algorithm for the order dispatch problem. 
    The key ingredients of our method lay in appropriately constructing a concordance function and applying the concordance penalty to the value-based objective function. 
    % \item \textcolor{red}{to discuss about how to add the theory}
    % pattern transferring in
    \item Empirically, we evaluate different algorithms in a well-calibrated simulator.  The proposed method achieves superior performance and demonstrates the usefulness of pattern transfer. 
\end{itemize}

\vspace{-.5cm}
\section{Related Work} % Background and 
% \subsection{Order dispatch problem}
% important and 
\textbf{Order dispatching} is a longstanding topic in the literature on intelligent transportation systems. 
Traditional methods typically only consider short-term performance.  
For example, \citet{liao2003real} proposes to match the nearest driver to each order, \citet{zhang2016control} uses the queue data structure to dispatch with the first-come-first-serve strategy, and \citet{zhang2017taxi} aims to optimize the success rate of the order matches by matching driver-order pairs within a short time window. % via combinatorial optimization.

However, these methods might lead to a sub-optimal policy in the long run, because they do not consider the long-term spatial equilibrium between the orders and drivers. 
% possibility that the potential orders in the future are more suitable for the drivers.
\citet{xu2018} made an important step forward by modeling the order dispatch problem as a Markov Decision Process (MDP) and designing a value-based RL algorithm, which can optimize resource allocation in a farsighted view and has shown great success in applications. 
% which has shown great success in applications. 
Further extensions have been proposed in the literature. 
For example, Li et al. \citep{li2019efficient} extended the single-agent setting in \cite{xu2018} to a  multi-agent setting, which is more capable of  modeling the complex interactions between drivers and orders.
Tang et al. \citep{tang2019deep} extended the tabular-like value function in \cite{xu2018} to deep value networks. 
% , which can improve the efficiency of the value iteration step.
% such as Deep Q-network based methods \citep{tang2019deep} or multi-agent RL \citep{li2019efficient}. 
Our work builds on this line of research and considers the  non-stationary nature of the order dispatch problem. 
\textbf{Transfer RL} aims to boost the training process in a \textit{target environment} by leveraging and transferring external knowledge from one or multiple \textit{source environments} \citep{zhu2020transfer}. 
% , which may be utilized to address the non-stationary problem in order dispatch problem. % \citep{williams1993tight, bertsekas2011approximate, rusu2015policy, zhu2020transfer}. 
Existing transfer RL approaches can be roughly categorized into reward shaping \citep{williams1993tight,  devlin2012dynamic}, learning from demonstrations \citep{bertsekas2011approximate, kim2013learning}, policy transfer \citep{rusu2015policy, czarnecki2019distilling},  inter-task mapping \citep{gupta2017learning, torrey2005using},  representation reuse  \citep{rusu2016progressive}, and learning disentangled representation \citep{dayan1993improving,  schaul2015universal}. 
See \citet{zhu2020transfer} for a recent survey. 

% In order dispatch problem, the real-time environment is usually unknown, and it is challenging for the agent to optimize the policy before a large amount of data have been collected. 

However, research on transfer RL in the order dispatch problem is limited. 
To the best of our knowledge, the only existing method is proposed by \cite{wang2018deep}, where several representation resue-type methods are adapted  to transfer the pre-trained neural network model in the source environment to the target environment. 
Such an approach requires the value function to be modeled by a deep neural network so as to share the weights of hidden layers. 
Therefore, it is not directly applicable to other settings, such as the tabular setting as in \citep{xu2018}. 
Our approach, on the contrary, is generally applicable to value function parameterized by any function class. 
% of any formats. 
% because the knowledge is transferring via penalty function in the value iteration step.
In addition, our method is more interpretable in the transferring process. 
Specifically, in this work, we focus on the concordance relationship, which yields intuitive explanations and provides additional insights. 
% is an important pattern, 
% to share between source and target environments in order dispatching. 
% Therefore, a penalty for concordance is adopted in the value iteration step.
Finally, our penalty-based approach can be easily extended when other kinds of domain knowledge about the connection between the source and the target  environment are available, and it is of separate interest.

% To the best of our knowledge, this is the first work considering transferring the structural pattern of an existing value function via a penalty function. 

% \noindent\textbf{Definition (Transfer RL):} Given a set of source MDPs $\mathcal{M}_s$ and a target MDP  $\mathcal{M}_t$, transfer RL aims to learn an optimal policy $\pi^*$ for the target MDP, by leveraging exterior information $\mathcal{D}_s$ from $\mathcal{M}_s$ as well as interior information $\mathcal{D}_t$ from $\mathcal{M}_t$:
% \begin{equation*}
%     \begin{split}
%         \pi^*  &= arg\max_{\pi} \mathbb{E}_{s \sim \tau^t, a \sim \pi}[Q_{\mathcal{M}_t}^\pi(s,a)] \\
%         & where\ \ \pi = \phi(\mathcal{D}_s, \mathcal{D}_t),
%     \end{split}
% \end{equation*}
% where $\tau^t$ denotes the observed states in target MDP. 

% In real-world RL applications, the real environment typically is unknown, and it is challenging for the agent to optimize the policy before a large amount of data have been collected. 

% what knowledge has been transfer --> value pattern

% How to transfer --> concordance

% 1. value-based RL
% 2. transfer RL
% 3. value stabalization / penalty in value estimation

% shaped rewards (Section 4.1), 
% expert demonstrations (Section 4.2), 
% teacher policies (Section 4.3). We also organize some TL approaches by the way transfer happens, such as inter-task mapping (Section 4.4), learning transferrable representations (Section 4.5 and 4.6), and etc.

\section{Setup}
% formally define the order dispatch problem
% In this section, we briefly review the value-based order dispatch framework proposed in \citet{xu2018}. 
% It is widely adopted in the literature \citep{qin2020ride, wang2018deep, tang2019deep} and will serve as a foundation of our approach. 
% Similar frameworks have been widely studied in the literature
% KDD solutions
% Semi-Markov decision process

Order dispatch problem can be formalized as a semi-Markov decision process (SMDP) model \citep{sutton1999between}, which is an extension of the MDP model in that SMDP allows the actions to be temporally extended. 
Various RL algorithms based on SMDP models have achieved great success in  real-world order dispatch applications \citep{xu2018, wang2018deep, tang2019deep, qin2020ride}.
Formally, the components of the SMDP can be built as follows:

\textbf{State.} We consider an episodic setting with each day as one episode.
The time and locations are discretized as $T$ time points and $N$ hexagon grids, respectively. 
% We consider a sequence of discrete time points $\mathcal{T} \equiv \{0, 1, \dots, T\}$ and a discrete set of regions $G = \{1, \dots, N\}$. 
The state space is $\mathcal{S} = \mathcal{T} \times G$, where $\mathcal{T} \equiv \{0, 1, \dots, T\}$ and $G \equiv \{1, \dots, N\}$.
At each time $t \in \mathcal{T}$, the state of a driver is a temporal-spatial pair $s = (t, i)$, where $i \in G$ is the driver's location.

\textbf{Action.} An available driver can either be assigned to serve an order or to stay idle. 

\textbf{Reward.} If one driver accepts an order with revenue $R$ and the order takes time $\Delta t$, then we consider the reward $R_{\gamma} = \sum_{t=0}^{\Delta t} \gamma^t \frac{R}{\Delta t}$, where $\gamma$ is the discount factor. 
If the driver chooses to stay idle, the reward is 0. 

\textbf{State transition.} 
For an available driver at state $s = (t,i)$, if the action is to serve an order, the driver will transit to state $s' = (t + \Delta t, i')$, where $\Delta t$ is the time cost and $i'$ is the destination; otherwise, the driver will keep idling and transit to state $s' = (t+1, i)$. 

\textbf{Value function.} 
We use $V_{\pi}(s)$ to denote the expected discounted cumulative reward that one random driver can collect starting from state $s$ to the end of the day, suppose the dispatch system follows the policy $\pi$. 
To simplify the notation, we may drop the subscript $\pi$, and use $V_{t,i}$ to denote $V(\small(t, i\small))$. 

\textbf{Policy.} 
The dispatching problem is a cooperative multi-agent RL task. 
At each decision point, the central agent will receive a list of order requests. 
A policy $\pi$ will assign these orders to idle drivers. 
The goal is to learn an optimal policy that maximizes the expected long-term reward.  

% Each independent driver can be modeled as an agent in RL settting. 
% Under the local-view setting

% balance the taxi supply and passenger demand across different areas of the city, so as to meet more passengers’ requests and reduce drivers’ vacant time.

% its spatial location indexed according to the provided grid dataset. 
% $(t_k, g_{it})$, where $t_k$ denotes the time-of-the-day discretized into 10-minute grids, and $g_{it}$ denotes its spatial location indexed according to the provided grid dataset.

% Therefore, both the transition and reward are deterministic, and the randomness lies in the future demands (the available actions at each state). 
% strictly speaking, our MDP is one with stochastic action sets (SAS-MDP)

% some graphs, diagrams might be needed here and above, to explain the system and the problem. 
% multi-agent
% via Concordance Relationship
% Value Transfer  for Order Dispatch
\section{Pattern Transfer Learning in Order Dispatch}
\subsection{Value-based order dispatch} 
With such an SMDP model, we can design a generalized policy iteration (GPI) approach   \citep{sutton2018reinforcement} to optimize the long-term cumulative reward. 
A GPI framework alternates between a policy evaluation step where we evaluate the value of each state, and a policy improvement step where we behave greedily with respect to the value so as to improve the current policy. 
The framework is summarized in Algorithm \ref{algotable: GPI}, with two key components detailed below. 
We note that similar frameworks have been considered in the literature \citep{xu2018,  tang2019deep, qin2020ride}, % wang2018deep, 
and the main differences lay in the details of the key components. 
In this work, a value transfer approach is designed to utilize existing offline data to evaluate the policy value more efficiently.
% make the policy evaluation step more efficient. 
% (Algorithm 2 in \citet{qin2020ride})
% The above algorithm is a one-step improvement over the behaviour policy in the provided dataset. 
% To keep improving our policy via value iteration

\begin{algorithm}
  \caption{Generalized Policy Iteration for Order Dispatching}
  \begin{algorithmic}[1]
  \STATE \textbf{Data:} transition buffer $\mathcal{D}$.
     \FOR{day $1, 2, \dots$}
     	\STATE Learn a value function $\widehat{V}$ from the data in $\mathcal{D}$ using a policy evaluation method (e.g., \eqref{formula:objective} or \eqref{eqn:DP_value}). % for example, the one in XX or the one to be proposed
     	\STATE Within each dispatch window throughout the day, match orders and drivers 
     	in a collectively greedy way with respect to  $\widehat{V}$ by solving \eqref{eqn:KM}. 
     	\STATE Add the new transition tuples into $\mathcal{D}$. 
	 \ENDFOR 
  \end{algorithmic}
  \label{algotable: GPI}
\end{algorithm}
% how to describe pi and pi_drive? 
% do we need trajectory data?

\textbf{Policy evaluation.}
At the beginning of each day, a data buffer of transitions tuples $\mathcal{D} = \{(s_j, a_j, r_j, s'_j)\}$ has been collected in previous days, where $s_j$ is the initial state, $a_j$ is the observed action, $r_j$ is the received reward, and $s'_j$ it the finish state. 
% $r_j$ is the discounted reward when $a_j$ is not to idle. 
% , s \in \mathcal{S}
We need to evaluate the value of each state $V(s)$ using the collected data. 
Various methods have been proposed, such as dynamic programming  \citep{xu2018} and deep-Q network \citep{tang2019deep}. 
We will detail our procedure in Section \ref{sec:eval_concordance}. 
% temporal difference \citep{xu2018}, 

\textbf{Online dispatch (policy improvement).}
% add some multi-agent; confilct, etc, here
% this goal can also be interpreted as finding the best action for each driver to optimize the future global gain in a coor- dinated way. 
Within each dispatch window, we need to match active orders and available drivers with the objective of maximizing the long-term collective cumulative rewards.
We will act in a greedy way with respect to the estimated value function $\widehat{V}$. 
Specifically, as a common procedure in the literature \citep{xu2018, tang2019deep}, we consider a bipartite matching problem for every possible driver-order pair: 
\begin{equation}\label{eqn:KM}
    \begin{split}
    \argmax_{\{a_{lk}\} \in C} \sum_{l=0}^m \sum_{k=0}^n \widehat{Q}(l,k) a_{lk}, 
    \end{split}
\end{equation}
where $l \in \{1,\dots, m\}$ corresponds to all available drivers, $k \in \{1,\dots, n\}$ corresponds to the active orders, and $a_{lk} \in \{0, 1\}$ is the indicator of assigning order $k$ to driver $l$ with $l = 0$ or $k = 0$ denoting no match. 
% with $i = 0$ denoting no-answer and $j = 1$ denoting keeping idling. 
Here, $C$ contains constraints including (i) $\sum_{l=0}^m a_{lk} = 1, \forall k$, indicating that each order can be assigned to at most one driver, (ii) $\sum_{k=0}^n a_{lk} = 1, \forall l$, meaning that each driver can take to at most one order, and (iii) some other business constraints.
The Q-function can be derived as $\widehat{Q}(l, k) = \gamma^{\Delta t(l, k)} \widehat{V}_{i'_k, t'_{lk}} + r_k$, where $\Delta t(l, k)$ is the time cost, $i'_k$ is the finish location, $t'_{lk}$ is the finish time, $r_k$ is the reward of order $k$. 
It is easy to verify that $\widehat{Q}(l, 0) =  \widehat{V}(s_l)$, where $s_l$ is the current state of driver $l$. 
% When $j = 0$, we have $\Delta t(i, j) = 0$
The Kuhn-Munkres (KM) algorithm \citep{munkres1957algorithms} can be applied to solve \eqref{eqn:KM}, 
and the advantage function trick in \citet{xu2018} can be used to reduce the computational cost.

% sample (predicted) advantage 
% refer to SMDP part?

% Refer to formula (6) and (7) in \cite{xu2018large} and related discussions therein for more details. 
% collective-
% common practice

\subsection{Concordance relationship}
As discussed in Section 1, we aim to utilize the pattern similarity between the old (source) environment and the current (target) environment through a concordance relationship between their value functions. 
% one major challenge to the value-based order dispatch algorithm is 
% to accurately estimate the value functions with limited data. 
%  at the policy evaluation step
% data scarcity! non-stationarity! 
% \hl{why standard non-stationary RL algorithm will fail?}
% Although there are usually lots of logged historical data, 
% given the non-stationarity, complex spatial-temporal dependency, and the  multi-agent nature of this problem, 
% it is not clear how to appropriately utilize them. 
% , due to the complex spatial-temporal structure of the system and the . 
% data generated in some old environments
% Instead, we propose to capture the pattern similarity  between the old environment and the current environment through a concordance relationship. 
% between some value function in the old environment and the values in the current environment. 
% no need to be optimal. But stable? lots of data? [clarify]
More precisely, suppose we have a value function $V^{s}$ learned by previous interactions with an source SMDP environment $\mathcal{M}^{s}$.  
$V^{s}$ could be learned by running Algorithm \ref{algotable: GPI} or other RL algorithms in this environment, or estimated from logged data. 
Our goal is to run Algorithm \ref{algotable: GPI} in the target environment $\mathcal{M}^{t}$ with the objective of maximizing the cumulative rewards.
% Our objective is to learn an optimal policy $\pi^*$ by running Algorithm  \eqref{algotable: GPI} in the current environment $\mathcal{M}^{t}$. 
We assume $\mathcal{M}^{t}$ and $\mathcal{M}^{s}$ share the same state space, rewarding system, and discount factor. 
However, due to environment non-stationarity, 
the spatial-temporal distribution of orders and drivers may change significantly, and so does the value function. 
Therefore, directly using the old dataset or value functions may cause huge bias. 

% transition probability distribution and the reward distribution may vary a lot.   
% \blue{some modifications required here.}
%  and action space $\mathcal{A}$, and we use the same 

% \textbf{some issues here. determinstic transition but random action set. Not a standard SMDP}
% Denote $\mathcal{M}_i = (\mathcal{S}, \mathcal{A}, \gamma, \mathcal{R}_i, \mathcal{T}_i)$.  

% the estimate the value of a new policy $\pi^{t}$ in a new environment $\mathcal{M}^{t}$. 
% Recall our challenges that 
% challenge with limited data

% \hl{some organization needs here}
% \hl{$V^*$ or any reasonable V_new?}

% be the same in both $\mathcal{D}^{s}$ and  $\mathcal{D}_{cur}$. 
% Let $E = \bigcup_i E$

% [Intuitions. Motivations]

% (s1, s2) \in E, \\
% For a set of user-specified state pairs $E = \{(s_1, s_2)\}$ on which we believe the concordance relationship should hold with a large probability, 

% \hl{cite some concordance paper. Ours} 
Motivated by the observations in Figure \ref{fig:data} that the relative relationship between the value of the hot regions and of the cold regions (or that between the rush hours and the normal hours) will be relatively consistent, we aim to capture this structural stability so as to transfer knowledge from the old data to stabilize the value estimation. 
Formally, 
the \textit{concordance relationship} on a state pair $(s_1, s_2)$ holds between two value functions $V$ and $V'$ if and only if
\begin{align*}
    [V(s_1) - V(s_2)] [V'(s_1) - V'(s_2)] \ge 0. 
\end{align*}
% This relationship implies that, if the value of state $s_1$ is higher than that of $s_2$ in $V$, then the same holds in $V'$, and vice versa. 
Given some pre-specified or estimated distribution $\mu$ over the space of state pairs $\mathcal{S} \times \mathcal{S}$, 
we define the \textit{concordance loss} between two value functions $V$ and $V'$ as 
% {\small 
\begin{align*}
    c(V, V'; \mu) \equiv \mathbb{E}_{(s_1, s_2) \sim \mu} &\mathbb{I}
\big\{[V(s_1) - V(s_2)] \\
&\times [V'(s_1) - V'(s_2)] < 0  %\\
\big\},
\end{align*}
% }
where $\I(\cdot)$ denotes indicator function. 
% {\small 
% \begin{align*}
% c(V, V'; \mu) \equiv \mathbb{E}_{(s_1, s_2) \sim \mu} \mathbb{I}
% \big\{[V(s_1) - V(s_2)] [V'(s_1) < V'(s_2)] < 0 %\\
% \big\}. 
% \end{align*}
% }
Here, $c(V, V';\mu)$ is the probability that the concordance relationship between $V$ and $V'$ will be violated, evaluated on $\mu$. 
The concordance function has been widely employed in applications such as classification tasks  \citep{cortes1995support} and optimal decision making \citep{liang2017sparse, fan2017concordance, shi2021concordance}. 
However, to the best of our knowledge, it is used in RL
for the first time. 

% $V^{t}$ and $V^{s}$ 
% evaluated on $\mu$ and $E$
% between $V^*$ and $V^{s}$ 
% of $\pi^*$ 
Let the optimal value function in the target environment be $V^*$. 
Motivated by the discussions above, we make the following assumption throughout this paper: 
the concordance relationship between $V^*$ and $V^{s}$ will hold with high probability, as evaluated on some appropriately chosen distribution $\mu$. 
More precisely, $c(V^*, V^{s}; \mu) \in [0, 1)$ is small. 
% \begin{asmp}\label{asmp:concordance_V}
% \textbf{(Concordance Relationship)}
% % For every $j \in [1,\dots,J]$, 
% More precisely, there is a small constant $\epsilon \in [0, 1)$, such that $c(V^*, V^{s}; \mu) \le \epsilon$. 
% % , \blue{for some distribution $\mu$ on the state pair and some pre-specified set E}. 
% % Where E is a pre-specified set for which we believe the concordance relationship will hold. 
% %  and $E$
% % , \mu
% % \begin{equation*}
% %   c(V^*, V^{s}; \mu) \le \epsilon. 
% % \end{equation*}
% \end{asmp}

In practice, for each state pair $(s_1, s_2)$, $\mu(s_1, s_2)$ should incorporate important domain knowledge about the importance of this pair as well as our belief that the concordance relationship on this pair will hold between $V^*$ and $V^{s}$. 
% , and different  
% \hl{some examples about mu. can also be combined! more data, more stable, less probability to be violated? , according to domain knowledge}
As an example, in this paper, we focus on the value concordance relationship between hot regions and cold regions. 
% We consider a simple $\mu$. 
Specifically, let $E$ be a set of user-specified location pairs on which the concordance relationship is believed to hold, 
we define 
\begin{align*}
    \mu(s_1, s_2) = (T|E|)^{-1}\I[(g(s_1), g(s_2)) \in E, t(s_1) = t(s_2)], 
\end{align*}
where $|E|$ is the cardinality of $E$, and $g(s)$ and $t(s)$ is the location and time component of the state $s$, respectively.

\subsection{Policy evaluation with concordance penalty}\label{sec:eval_concordance}
% Value estimation
% \begin{equation*}
% \begin{split}
%     \widehat{Q}^{t} = argmin_Q \; \frac{1}{n}\sum_i [\widehat{\R}^{\pi^{t}} Q(s_i, a_i)]^2
%     + \sum \Big\{\lambda * c(Q, \widehat{Q}; E, \mu) \Big\}
% \end{split}
% \end{equation*}

In this section, we discuss how to improve the efficiency of value function estimation by utilizing the dataset collected in the source environment through a concordance penalty function. 
To simplify the notation, for every $t$, we denote $\vV_t = \{V_{t,i}\}_{i=1}^N$. 
We similarly define $\vV^{s}_t$ and $\widehat{\vV}_t$. % and $\vV^*_t$. 

A straightforward approach to estimate the value function is dynamic programming (DP) \citep{xu2018}.
Let $\widehat{V}_{T, i} = 0$ for every $i \in G$.
For $t = T - 1, T - 2, \dots, 0$, $\widehat{V}_{t, i}$ for every $i \in G$ is calculated as 
\begin{equation}\label{eqn:DP_value}
    \widehat{V}_{t, i} = \frac{1}{|\mathcal{D}(t, i)|} \sum_{j \in \mathcal{D}(t, i)}( \gamma^{\Delta t(a_j)} \widehat{V}_{t'_j, i'_j } + r_j), 
    % R_{\gamma}(a_j)
\end{equation}
where $s'_j = (i'_j, t'_j)$ and $\mathcal{D}( t, i) = \{j : s_j = (t,i)\}$ denoting tuples with current state $(t,i)$.

To present our method, we note that the DP-based policy evaluation step \eqref{eqn:DP_value} is equivalent to minimizing the squared temporal-difference (TD) error \citep{sutton2018reinforcement}. 
% any issues with this derivation? 
% Denote $\widehat{\vV}_t = \{\widehat{V}_{i,t}\}$. 
Specifically, %at a specific policy evaluation time point, with $\mathcal{D}$ be the set of transition tuples collected from the current environment, 
% can be updated with more online collected data. 
we first set $\widehat{V}_{t,i} = 0$ for every $i$, and then solve the following optimization problem recursively, for $t = T - 1, T - 2, \dots, 0$: 
\begin{equation}\label{formula: TD}
    \widehat{\vV}_t = \argmin_{\vV_t} 
    \sum_{j \in \mathcal{D}(t)} [V_{t, i_j} - \gamma^{\Delta t(a_j)} \widehat{V}_{t'_j, i'_j} - R_{\gamma}(a_j)]^2. 
    %  \sum_{j \in \mathcal{D}( t)} (V_{i,t} - \gamma^{\Delta t(a)} \widehat{V}_{i', t'} - R_{\gamma}(a))^2
\end{equation}
It is easy to verify that, the estimated value function $\widehat{\vV}$ by solving \eqref{formula: TD} is the same with the output of \eqref{eqn:DP_value}. 
% Algorithm XX. 

% $\widehat{V}_{t'_j, i'_j} \I(t'_j > t) + V_{t, i_j} \I(t'_j = t)$
% under Assumption \ref{asmp:concordance_V}, 
With such an observation, we propose to estimate the value function by minimizing the squared TD error with the concordance constraint. 
For any time index $t$ and any two value functions $V$ and $V'$, we define the \textit{spatial concordance loss} between $\{V_{t,i}\}_{i=1}^N$ and $\{V'_{t,i}\}_{i=1}^N$ as 
% $ l(\{V_{t,i}\}_{i=1}^N, \{V'_{t,i}\}_{i=1}^N; E) \equiv  
%     \sum_{(i,j) \in E}
%      \mathbb{I}\Big\{[V_{t,i} - V_{t,j}][V'_{t,i} - V'_{t,j}] < 0
%      \Big\}. $
% Here, $l(\vV^*_t, \vV^{s}_t; E)$ is , defined as 
% $\vV^*$ and  $\vV^{s}$ becomes
% {\small
\begin{equation*}
    \begin{split}
l(\{V_{t,i}\}_{i=1}^N, \{V'_{t,i}\}_{i=1}^N; E) 
&\equiv  
    \sum_{(i,j) \in E}
     \mathbb{I}\Big\{[V_{t,i} - V_{t,j}]\\
     &\quad \times [V'_{t,i} - V'_{t,j}] < 0
     \Big\}. 
     \end{split}
\end{equation*}
% }
Then, let $\mathcal{D}(t) = \bigcup_i \mathcal{D}( t, i)$, we can obtain $\widehat{\vV}_t$ by solving
\begin{equation}\label{formula:TD_constraint}
    \begin{split}
    &\argmin_{\vV_t} 
    \sum_{j \in \mathcal{D}(t)} \Big[V_{t, i_j} - \gamma^{\Delta t(a_j)} \widehat{V}_{t'_j, i'_j} - R_{\gamma}(a_j)\Big]^2\\
    &\;\; s.t. \;\;\;\;\;\;\;\; l(\vV_t, \vV^{s}_t; E) \le \epsilon.% |E|. 
    % \sum_{(i,j) \in E}\;
    %  \mathbb{I}\{(V_{i,t} > V_{j,t})(V^{s}_{i,t} < V^{s}_{j,t}) < 0
    %  \} \le \epsilon |E|
    %  \sum_{j \in \mathcal{D}( t)} (V_{i,t} - \gamma^{\Delta t(a)} \widehat{V}_{i', t'} - R_{\gamma}(a))^2
    \end{split}
\end{equation}
To solve this constrained optimization problem, a equivalent penalized optimization problem is considered: 
\begin{equation}\label{formula:TD_penalty}
    \begin{split}
    \argmin_{\vV_t} \Big\{
    &\sum_{j \in \mathcal{D}(t)} \Big[V_{t, i_j} - \gamma^{\Delta t(a_j)} \widehat{V}_{t'_j, i'_j} - R_{\gamma}(a_j)\Big]^2 \\
    &+ \lambda \times l(\vV_t, \vV^{s}_t; E) \Big\}, 
    \end{split}
\end{equation}
where $\lambda > 0$ is the Lagrange parameter. 
By the Lagrange duality, we know that, for any $\epsilon > 0$, there exists some $\lambda > 0$ such that the solution of \eqref{formula:TD_penalty} is the same with that of \eqref{formula:TD_constraint}. 
% note that with 

% , and commonly used in the literature . 
% convex relaxation. 
Finally, we note that problem \eqref{formula:TD_penalty} is not differentiable. 
In practice, the hinge loss, which is a convex upper bound of the concordance loss function, has been commonly used as a surrogate loss function \citep{liang2017sparse, cortes1995support}. 
Specifically, the hinge loss between $\vV_t$ and  $\vV^{s}_t$, can be written as 
% $h(\vV_t, \vV^{s}_t; E) = \sum_{(i,j) \in E}\Big\{
%      \I[V^{s}_{t,i} < V^{s}_{t,j}]
%      [1 - (V_{t,j} - V_{t,i})]_{+} + \I[V^{s}_{t,i} > V^{s}_{t,j}]
%      [1 - (V_{t,i} - V_{t,j})]_{+} \Big\}$. 
\begin{align*}
h(\vV_t, \vV^{s}_t; E) &= \sum_{(i,j) \in E}\Big\{
     \I[V^{s}_{t,i} < V^{s}_{t,j}]
     [1 - (V_{t,j} - V_{t,i})]_{+} \\
     &\;\;\;\;\;\;\;\;\;\;+ \I[V^{s}_{t,i} > V^{s}_{t,j}]
     [1 - (V_{t,i} - V_{t,j})]_{+} \Big\}
\end{align*}

% , denoted as $h(\vV_t, \vV^{s}_t; E)$

% Plugging (\ref{formula:hinge_loss}) into (\ref{formula:original_optimization}), we get the optimization objective function $\mathcal{L}(\{V_{i,t}\}_{i=1}^N;
% \{V_{i,t}^{s}\}_{i=1}^N, \lambda)$ as 
% \mathcal{D}
% \hl{After finishing the preliminary part. Connect with here.}
% in Algorithm XX
Putting all the discussions together, we propose to replace the DP-based policy evaluation step 
\eqref{eqn:DP_value} by solving the following optimization problem recursively, for $t = T - 1, \dots, 0$: 
\begin{eqnarray}\label{formula:objective}
    \begin{split}
    \widehat{\vV}_t = \argmin_{\vV_t} \bigg\{
    &\sum_{j \in \mathcal{D}(t)} \Big[V_{t, i_j} - \gamma^{\Delta t(a_j)} \widehat{V}_{t'_j, i'_j} - R_{\gamma}(a_j)\Big]^2 \\
    &+ \lambda \times h(\vV_t, \vV^{s}_t; E) \bigg\}. 
    \end{split}
\end{eqnarray}

\textbf{Optimization.}
Let the objective function of \eqref{formula:objective} be $\mathcal{L}( \vV_t; \vV^{s}_t, \mathcal{D}(t), \lambda)$. 
For every $i = 1, \dots, N$, we can derive the partial gradient as 
\begin{equation}\label{formula:gradient}
    \begin{split}
  &\frac{\partial}{\partial V_{i,t}}  
  \mathcal{L}( \vV_t; \vV^{s}_t, \mathcal{D}(t), \lambda) \\
    % &= 
    % 2 \sum_{j \in \mathcal{D}_{new}(i, t)} (V_{i,t} - \gamma^{\Delta t(a)} \widehat{V}_{i', t'} - R_{\gamma}(a)) \\
    %  &+ 
    % \frac{\partial}{\partial V_{i,t}} \lambda \left[\sum_{(i,j) \in E: V^{s}_{i,t} < V^{s}_{j,t}}
    %  [1 - (V_{j,t} - V_{i,t})]_{+}
    %  + \sum_{(i,j) \in E: V^{s}_{i,t} > V^{s}_{j,t}}
    %  [1 - (V_{i,t} - V_{j,t})]_{+}\right]\\
    &= 
    2 \sum_{j \in \mathcal{D}(i, t)} (V_{t,i_j} - \gamma^{\Delta t(a_j)} \widehat{V}_{t'_j, i'_j} - R_{\gamma}(a_j)) \\
     &- 
    \lambda
    \sum_{j: (i,j) \in E}
    \big[ \mathbb{I}(V^{s}_{t,i} < V^{s}_{t,j}, V_{t,j} - V_{t,i} < 1)\\
     &+ \mathbb{I}(V^{s}_{t,i} > V^{s}_{t,j}, V_{t,i} - V_{t,j} < 1). 
     \big]
     \end{split}
\end{equation}
The explicit form of the gradient $\frac{\partial}{\partial \vV_t}  
  \mathcal{L}( \vV_t; \vV^{s}_t, \mathcal{D}(t), \lambda)$ then follows. 
To solve \eqref{formula:objective}, we apply gradient descent with step sizes chosen by a diminishing step size rule \citep{boyd2003subgradient}. 
We have the following convergence guarantee. 
\begin{proposition}[Convergence]
With a diminishing step size rule, our gradient descent optimization algorithm will converge to the solution of \eqref{formula:objective}. 
\end{proposition}

\textit{Proof.}
% \begin{proof}
    For the objective function, both the loss part and the penalty part is a composition of a convex function and an affine function, and hence it is convex. 
    Besides, because the hinge loss is a subdifferentiable function, 
    it is easy to verify that the objective function is also a subdifferentiable function. 
    Therefore, according to \citet{boyd2003subgradient}, 
    a gradient descent optimization algorithm with a diminishing step size schedule for a convex subdifferentiable objective function will converge to the global optimum. 

\begin{figure*}[t!]
  \centering
\begin{subfigure}{\textwidth}
  \centering
  \includegraphics[width=\textwidth]{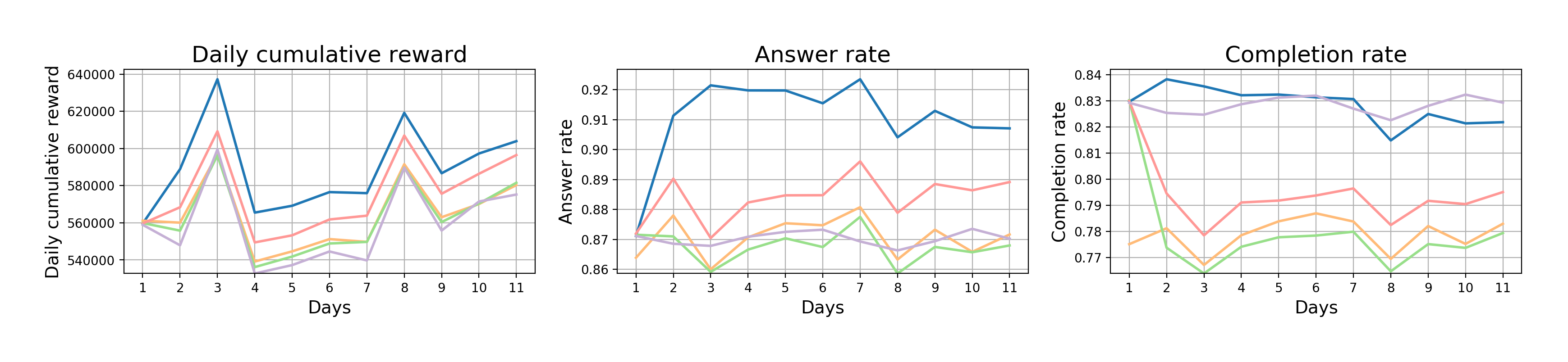}
\end{subfigure}\\
\begin{subfigure}{\textwidth}
  \centering
    \includegraphics[width=\textwidth]{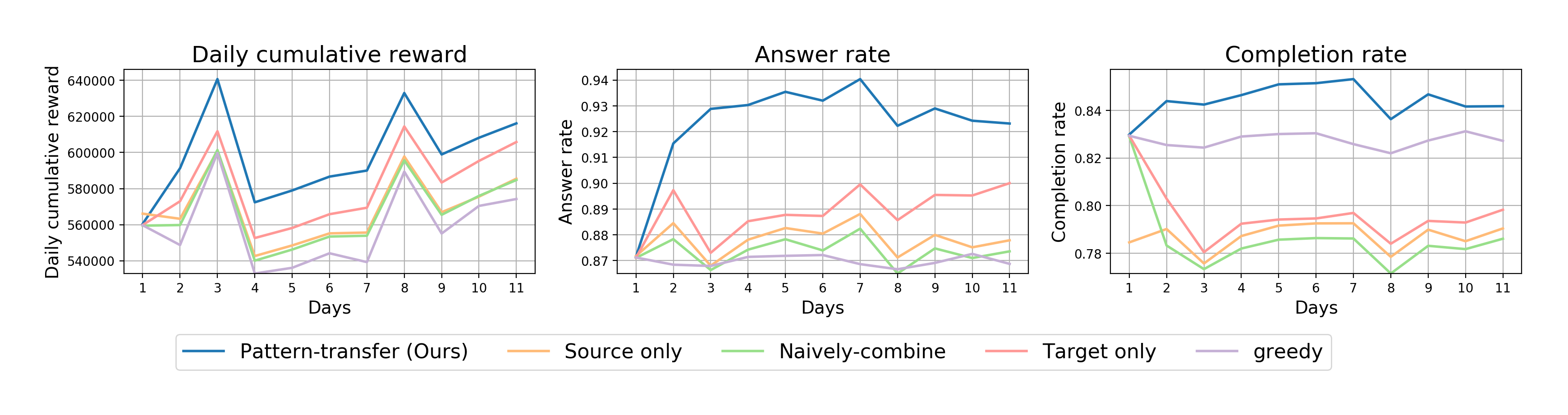}
\end{subfigure}%
 \caption{
Performance of different methods when $\gamma = 0.9$ (upper) and  $\gamma = 0.95$ (lower). 
The x-axis represents consecutive weekdays in the target environment. 
% The y-axis represents the different metrics to evaluate the performance for each day. 
Our method outperforms the baseline methods under different metrics.
}
 \label{fig:result}
\end{figure*}

\section{Experiments}\label{sec:exp}
% neigh or all pairs: neigh now
% 10 minutes as the length of one 

% 6.2 Simulator
% Beyond the toy example, we further evaluate the proposed MDP method on a more complicated and realistic dispatch simulator. In our application, offline simulator’s results are sometimes as impor- tant as online experiments considering that the online system is highly dynamic.
% Implementation Details. The simulator used here is built to perform a thorough modeling of the physical world for the on- demand ride-hailing platform. A typical sample implementation is to simulate a particular day from historical data. To do so, we first rewind real-happened orders in this date as demands. Each driver is initialized according to the location and time of his or her first entry in the platform. Drivers’ actions afterwards are totally determined by the simulator, either according to a user-defined order dispatch algorithm (serving orders), or using models fitted from historical data (idle movements and offline/online operations). The simulator is carefully calibrated, with the difference between simulated results and real-world metrics for the same day being within 2% in most cases, in terms of answer rate and total GMV.

\textbf{Simulator.}
To evaluate the proposed method, we build a real data calibrated dispatch simulator. 
Such simulator are constructed based on the open dataset from the DiDi ride-sharing platform \footnote{https://gaia.didichuxing.com}. 
This dataset contains drivers' trajectories, transition probability of idle drivers, information of order requests and hexagonized map grids in Chengdu, China, for 30 days.
% In the dataset, the distributions of trajectories and order requests are different among different days. 
Our simulator design follows the procedures introduced in \citet{xu2018}. 
Specifically, the order requests and drivers' online time periods are kept the same with the real data. 
After logging-in, the drivers will completely follow the dispatch algorithm. 
Other information such as the transition probability of idle drivers and the cancellation rates are all provided by or fitted from the data. 
The difference between the simulated results from our simulator and the official simulator is less than 5\% in terms of answer rate and total GMV.

\textbf{Setting.}
Following \citet{xu2018}, we use the first 15 days as the source environment and the latter 15 days as the target environment. 
In the real dataset, there exists a huge difference in the environment between weekdays and weekends. 
In this experiment, we focus on the weekdays only. 
Different value-based dispatch policies are run during $11$ weekdays in the latter half a month and their performance is recorded. 
All of these policies use the KM algorithm \eqref{eqn:KM} for dispatch, and the only difference lies in the choice of the $Q$-values. 
For policies relying on data generated in the target environment, we use the greedy policy as the initial policy on the first day. 
The following policies are considered: 
% (i) \textbf{Greedy (Myopia)}: only instant order rewards are considered. Replace $\widehat{Q}(i,j)$ in \eqref{eqn:KM} by $r_{ij}$; 
% (ii) \textbf{Distance-based}: only the distances are considered. Replace $\widehat{Q}(i,j)$ in \eqref{eqn:KM} by $-d_{ij}$; 
% (iii) \textbf{Source only}: the value functions are calculated using the source data only. Details can be found in \cite{xu2018}; 
% (iv)  \textbf{Target only}: the value function is initialized with zero and updated for the latter 15 days using TD updates ( Equation \ref{formula: TD}) with no penalty; 
% (v) \textbf{Naively-combine}: the value functions are calculated from the first 15 days and updated for the latter 15 days using TD updates with no penalty; 
% (vi) \textbf{Pattern-transfer}: our proposed concordance penalty based value evaluation methods.

\begin{itemize}
    \item \textbf{Greedy (Myopia)}: Only instant order rewards are considered. Replace $\widehat{Q}(l,k)$ in \eqref{eqn:KM} by $R_{lk}$. 
    % , i.e. the driver is intent to take the order with highest reward.
    % \item \textbf{Distance-based}: Only the distances are considered. Replace $\widehat{Q}(i,j)$ in \eqref{eqn:KM} by $-d_{ij}$.
    % The value is defined as the negative distance $A_{ij} = -distance(order_i, driver_j)$, i.e. the driver is intent to take the nearest order.
    \item \textbf{Source-only}: The value functions are calculated using the source data only. Details can be found in \cite{xu2018}.
    \item \textbf{Target-only}: The value function is initialized with zero and updated for the latter 15 days using TD updates ( Equation \ref{formula: TD}) with no penalty.
    \item \textbf{Naively-combine}: The value functions are calculated from the first 15 days and updated for the latter 15 days using TD updates with no penalty.
    \item \textbf{Pattern-transfer}: Our proposed concordance penalty-based value evaluation methods.
\end{itemize}

\begin{figure*}[t!]
  \centering
  \includegraphics[width=\textwidth]{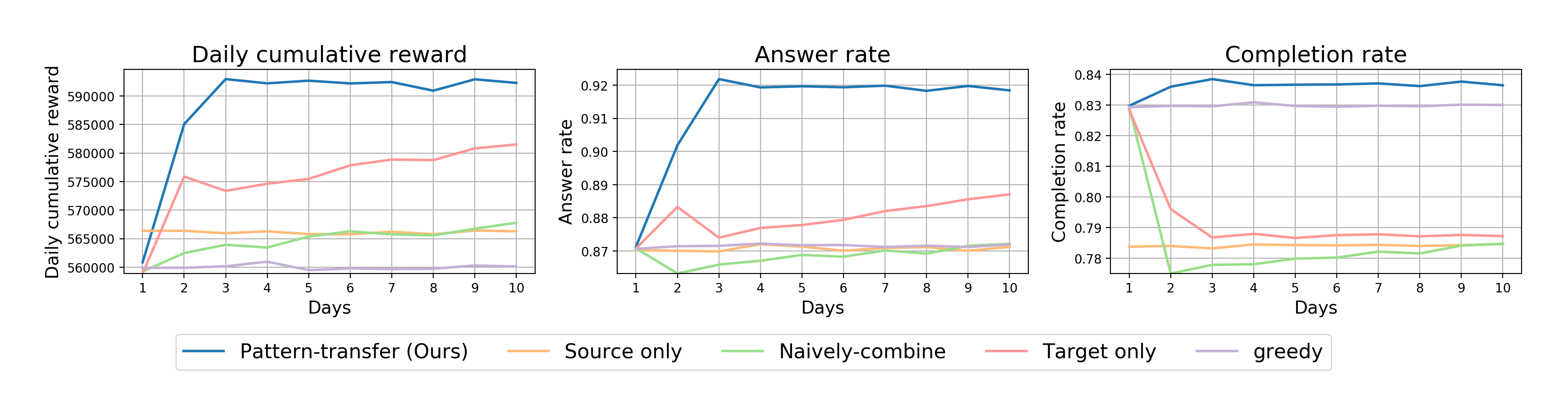}
 \caption{Results for different methods when the same day is repeatedly simulated for multiple times. 
 The x-axis represents repeated iterations of this single day in the target environment. 
%  The y-axis represents the different metrics. 
 Our method shows a stable performance within 2 iterations, while the target-only method requires more iterations to improve and  converge. 
%  , the proposed method (ours) outperforms the baselines methods in cumulative reward, answer rate and completion rate.
}
 \label{fig:repeat_one_day}
\end{figure*}
%  , the proposed method (ours) outperforms the baselines methods in cumulative reward, answer rate and completion rate.

\textbf{Results.}
The performance of different methods is summarized in Figure \ref{fig:result}. 
We consider two choices of $\gamma$,  $0.9$ and $0.95$. 
In addition to the cumulative rewards, we also report the answer rates (proportion of orders being answered) and the completion rates (proportion of accepted orders being eventually completed). 
Both metrics are commonly used to reflect the quality of a  dispatch algorithm \citep{xu2018}. 
We summarize our findings as follows. 
We focus on the cumulative reward as the main evaluation metric. 
% where we mainly comment on the cumulative rewards:
\begin{itemize}
    \item The proposed method outperforms the baselines by a significant margin. Specifically, compared with the target-only algorithm (i.e., no pattern transfer), the proposed method enjoys a jumpstart in the first several days, converges more quickly, and the convergence performance is better. 
    These improvements come from the efficient pattern transfer via the concordance penalty; 
    \item The performance of the target-only algorithm also improves slowly as the data accumulates, but the rate of improvement is much slower than the proposed method; 
    \item Both the source-only and the naively-combine algorithms suffer from the bias incurred by the non-stationarity of the environment; 
    \item The greedy policy does not consider the long-term performance, will cause undesired demand-supply distribution, and hence performs the worst. 
\end{itemize}

% Target cities will benefit from , which is the increased initial performance at the start; 2) Learning is more efficient, which means the convergence would happen earlier during training; 3) 

To see the convergence speed of different methods under the GPI framework more clearly, 
we conduct an extra experiment, where we repeat the simulation of a single day multiple times. 
Results are shown in Figure \ref{fig:repeat_one_day}.
The performance of our method typically converges within $2$ iterations and achieves superior performance, 
while the value of the target-only algorithm increases at a much slower rate. 
% while the performance of the target-only algorithm improves  much slower. 
This experiment further demonstrates the usefulness of the proposed pattern transfer method.

% We can also show by the accuracy of the value function estimation?
% concordance relationship. The magnitude? 

% Objective: in terms of (i) the values and (ii) the policy performance: 
% \begin{enumerate}
%     \item Faster than no offline data
%     \item Higher value (faster) than naively using offline data
% \end{enumerate}

% Performance & Convergence of Policy Iteration. In Section 5, we mentioned that the proposed learning and planning algorithm can be conducted in an iterative way in the principle of policy itera- tion. Here we evaluate this approach in the simulator by repeating the same day of simulation as a stable environment and perform- ing policy iteration on it. In particular, we set the distance-based method as the initialized policy and iteratively updates value func- tions and policy. Table 1 shows the results compared to the baseline distance-based method. The updated policy achieved a higher total GMV and completion rate compared to the baseline policy and the MDP policy without policy iteration. This process typically converges within ten iterations in terms of the performance gain.

\section{Discussion}

In this paper, we propose a novel pattern transfer method for the online order-dispatch problem. 
At the heart of our method is a concordance penalty, which efficiently captures the value patterns. 
Integrated with the GPI framework, the algorithm demonstrates superior performance in dealing with non-stationary environments. 

There are several future directions worthy of study. 
First, we currently model the value function without function approximation. It would be interesting to couple our proposal with some state-of-the-art universal function approximators, e.g., deep neural networks. 
Second, we can consider applying the pattern transfer learning method to more complex problems in intelligent transportation systems, such as multi-agent RL for order dispatch, joint order dispatching and fleet management, etc. 
Third, the concordance relationship is only one kind of pattern, and the idea of penalty-based transfer RL can be more general than the setup considered in this paper. 
% Other penalty functions and broader applications of this transfer RL framework can be considered. 
it is practically interesting to apply the proposed methodology to other domains to evaluate its empirical performance. 
Lastly, providing theoretical guarantees for the proposed method is also a meaningful next step. 

% deep neural network-based order dispatch, 

% Order dispatch is one of the central problems to ride sharing platforms. 
% Recently, value-based reinforcement learning algorithms have shown promising performance on this problem. 
% % under stationarity setting of environment.
% % With the development of Markov Decision Process (MDP) framework for the order-dispatch problem, existed reinforcement learning  based methods have achieved significant 
% However, in real-world applications, the non-stationarity of the demand-supply system poses challenges to re-utilizing the data generated in different time periods. 
% % Effectively transferring value patterns learned in a source environment to a target environment is critical to improving the performance of an order dispatch algorithm. 
% % Concordance, which measures the agreement between two variables, e.g. to evaluate hot zones in different environment, provides a good resource to address the challenge. 
% In this work, motivated by the fact that the relative value relationship between regions are relatively stable across various environments, we propose a pattern transfer framework for the order dispatch problem. 
% Our method efficiently captures the value patterns by incorporating a concordance penalty.
% The superior performance of the proposed method is supported by experiments. 

\appendix

\newpage

\section*{Acknowledgement}
We thank Didi Chuxing for sharing the datasets for public research.

\bibliographystyle{named}
\bibliography{main}

\end{document}